\documentclass[cameraready]{Interspeech}
\title{Leveraging Audio-LLMs to Filter Speech-to-Speech Training Data\thanks{\url{https://github.com/chin-alt/S2S-Filtering}.}}

\author[affiliation={1}]{Qixu}{Chen}
\author[affiliation={1,2}]{Satoshi}{Nakamura}
\address{
$^1$ School of Data Science, \\
$^2$ School of Artificial Intelligence \\
The Chinese University of Hong Kong, Shenzhen, China
}

\email{qixuchen@link.cuhk.edu.cn, snakamura@cuhk.edu.cn}


\usepackage{comment}

\begin{document}

\maketitle

\begin{abstract}
Large-scale mined corpora provide abundant training data for end-to-end speech-to-speech translation (S2ST) but may contain noise, misalignment, and semantic errors. Filtering noisy data is crucial to maintain robust speech translation performance. We study how to train an audio-language model to make keep/drop decisions on paired speech directly from audio. To obtain reliable supervision without manual labels, we adopt a scalable two-stage Rank→Distill strategy. A lightweight ranker generates keep/drop pseudo-labels from noisy speech pairs, then trains an audio large language model to predict keep/drop directly from raw paired speech. The resulting model jointly captures acoustic fidelity and cross-lingual semantic consistency for the selection of speech-conditioned data. Experiments on CVSS-C and SpeechMatrix show consistent improvements over unfiltered training, yielding up to +1.4 ASR-BLEU for end-to-end S2ST. 
\end{abstract}

\section{Introduction}
End-to-end speech-to-speech translation (S2ST) directly maps source
speech to target-language speech, reducing latency and avoiding error
propagation compared with cascaded ASR--MT pipelines
\cite{weiss2017,jia2019direct}.  Despite recent progress, S2ST remains
strongly constrained by the quality of available training data. In
practice, large-scale paired speech corpora are often mined or automatically
aligned, and therefore contain substantial noise, segmentation errors,
and inconsistent translation fidelity\cite{weiss2017}. These issues weaken supervision,
destabilize training, and make it difficult to identify reliable
speech--translation pairs for model learning.

Data filtering has been widely studied in text-based machine translation, including denoising with trusted data \cite{wang2018denoising}, curriculum-based selection \cite{zhang2019curriculum}, and embedding-based mining and filtering for noisy parallel corpora \cite{chaudhary2019low,guo2018effective}. However, end-to-end S2ST filtering remains underexplored and is challenging because both sides are speech.

% Only a limited number of studies have examined data filtering for end-to-end S2ST. For example, \cite{srivastava2024case} explores simple heuristics for mined speech corpora using ratio-based features, such as source--target duration ratios and source--target length ratios measured on ASR transcripts, together with model likelihoods. Speech-conditioned quality estimation methods such
% as BLASER 2.0-QE \cite{dale-costa-jussa-2024-blaser} also touch on data
% selection using learned semantic scores. While useful, these approaches
% are either heuristic-based or rely on single-modality signals, and thus
% often fail to capture acoustic degradations, semantic mismatches, and
% cross-modal inconsistencies in large mined ST datasets. Reliable
% quality assessment for mined ST data therefore remains a crucial problem.
Only a limited number of studies have explored data filtering for end-to-end S2ST. Prior work largely relies on simple heuristics, such as source--target duration ratios, transcript-length ratios from ASR, and model likelihoods \cite{srivastava2024case}. Speech-conditioned QE metrics such as BLASER 2.0-QE \cite{dale-costa-jussa-2024-blaser} provide learned scores that primarily reflect cross-lingual semantic alignment. While useful, these signals are either heuristic or semantics-dominated, and are often insensitive to S2ST-specific noise such as acoustic degradations, synthesis artifacts, and speech-pair misalignment in large mined corpora. Reliable quality assessment for mined S2ST data therefore remains a crucial problem.

LLM-based data filtering has recently emerged as an effective paradigm
for curating noisy parallel text, providing flexible quality judgments
beyond heuristic rules \cite{kocmi2023llmasjudge,waldendorf-etal-2025-multilingual}.
Such success naturally motivates extending the idea to speech translation. A simple strategy is to transcribe speech with ASR and apply text-based
LLM filtering on the transcripts. However, transcripts alone do not
fully reflect acoustic quality or speech-pair misalignment, limiting
the effectiveness of filtering. 

Recent advances in audio-language models suggest that such models can
directly perceive and evaluate speech quality from raw audio.
For instance, \cite{chen2025alld} demonstrates that audio LLMs can be
trained to generate descriptive quality assessments, indicating that they possess
end-to-end awareness of both acoustic and semantic properties.
This raises a key question: \emph{can audio-language
models be trained to decide which speech pairs to keep or discard?}
In practice, however, off-the-shelf
audio LLMs (e.g., Qwen2-Audio) often struggle to reliably follow
fine-grained filtering instructions, and large-scale speech pairs rarely come with reliable labels.

To address this, we adopt a two-stage \emph{Rank$\rightarrow$Distill}
self-bootstrapping strategy. We first learn
a coarse ranking under weak supervision and then distill high-confidence
extremes into an audio-language model trained to predict keep/drop
decisions directly from paired speech. This approach follows the
principle that when reliable labels are unavailable, learning a stable
ordering is often easier than learning absolute decisions.
This design is inspired by teacher--student self-training with pseudo
labels \cite{xie2020noisystudent} and prior evidence in speech that
high-quality pseudo labels can effectively support large-scale training \cite{hwang2022pseudo}. The resulting audio-language model can directly
assess paired speech by jointly considering acoustic fidelity and
cross-lingual semantic consistency.

The contributions of this paper are: 
(1) We propose a two-stage Rank$\rightarrow$Distill strategy to mine
reliable supervision from noisy speech translation data and train
audio-language models for filtering. 
(2) We show that audio-language models can perform effective
speech-conditioned data selection and improve S2ST training.

\section{Method}
\begin{figure*}[t]
\centering
\includegraphics[width=\textwidth]{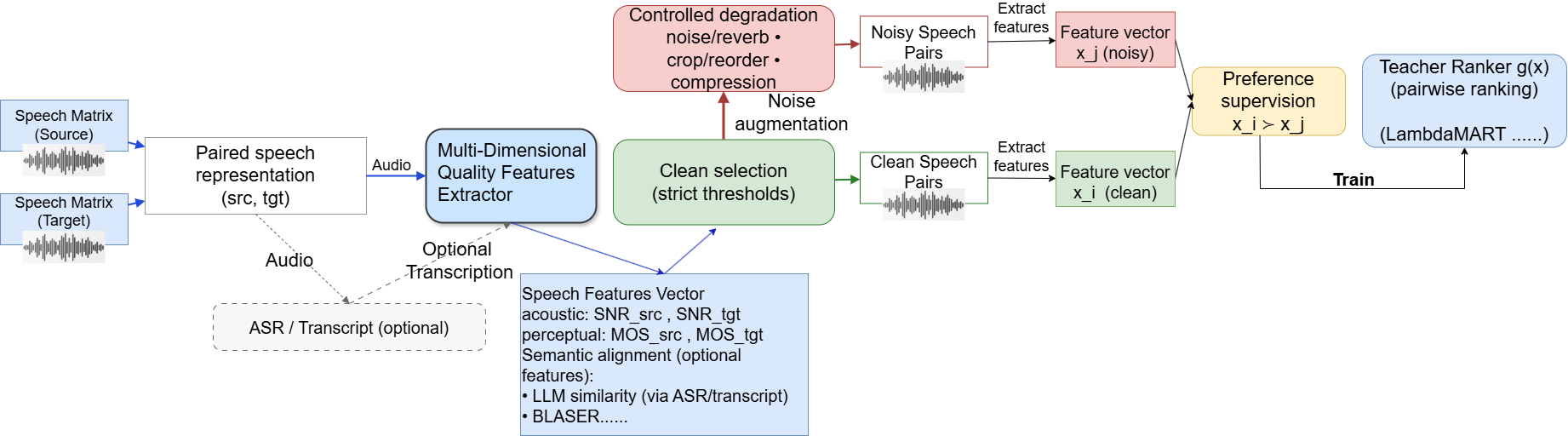}
\caption{
Construction of supervised training pairs and ranker training pipeline.
Clean speech pairs are selected using strict thresholds over acoustic,
perceptual, and semantic signals. Controlled degradations are applied to
generate negative examples. A lightweight teacher ranker is trained using
pairwise preference supervision between clean and noisy pairs.
}
\label{fig:ranker_training}
\end{figure*}
\subsection{Audio-LLM Filtering for S2ST Data Selection}

We study data filtering for speech-to-speech translation (S2ST) using an
audio-language model that takes paired source and target speech as input
and predicts a keep/drop decision. Unlike traditional pipelines that
combine hand-designed signals and thresholds, we treat filtering as a
direct decision problem over paired speech.

Let
\[
D=\{(x_i^{src},x_i^{tgt})\}_{i=1}^{N}
\]
denotes a large corpus of mined speech pairs containing varying levels of
acoustic noise and semantic mismatch. Our goal is to select a subset
$D_s \subset D$ that leads to better BLEU when used to train an S2ST
model, by retaining pairs that are both acoustically reliable and
semantically consistent.
\subsection{Dataset Generation}
\label{sec:dataset_generation}
\subsubsection{Quality signals for supervision.}
\label{sec:signals}
To construct supervision for training Audio LLM-based filters, we use a set of
automatically computed signals that approximate speech translation quality from
complementary perspectives. These signals are not used directly for filtering,
but to identify high-confidence training examples. We consider three aspects:
acoustic fidelity, perceptual quality, and semantic alignment.

\textbf{Acoustic fidelity supervision.}
For each speech signal, we compute the signal-to-noise ratio (SNR) as an
objective measure of background noise and recording conditions.
Samples with extremely low SNR are treated as unreliable for model learning.

\textbf{Perceptual quality supervision.}
In addition to physical noise measures, we estimate Mean Opinion Score (MOS) \cite{viswanathan2005mos}
to approximate human-perceived speech
naturalness. MOS captures perceptual degradations such as distortion, clipping, and unnatural prosody that may not be reflected in SNR alone.

\textbf{Semantic alignment.}
Semantic consistency supervision is derived from both text-mediated and
representation-based signals. Three variants are considered:
(i) LLM-based judgment over ASR transcriptions following an
\emph{LLM-as-a-judge} paradigm \cite{kocmi2023llmasjudge};
(ii) text-mediated comparison using machine translation outputs; and
(iii) representation-level agreement computed within a shared speech embedding
space. In practice, these signals are treated as optional and can be used
individually or in combination depending on computational budget.

\subsubsection{Construction of supervised training pairs.}
A development dataset is constructed to distinguish high-quality speech pairs
from noisy or unreliable ones, enabling the model to learn a decision boundary over speech translation quality. The overall process is illustrated in Figure~\ref{fig:ranker_training}.

\textbf{Clean pairs.}
High-confidence positive examples are selected by enforcing strict thresholds across acoustic, perceptual, and semantic dimensions.

\textbf{Noisy pairs augmentation.}
Negative examples are created by applying controlled perturbations to clean
speech pairs. We consider three types of degradation: (i) acoustic noise and
reverberation, (ii) mild temporal inconsistencies such as cropping or local
segment reordering that may disrupt cross-lingual alignment, and (iii) signal fidelity loss from compression artifacts. Perturbations are sampled from three presets
(\emph{light}, \emph{medium}, \emph{heavy}) with a 3:6:1 ratio,
covering realistic degradation levels in mined S2ST data.

To avoid trivial separation, we introduce controlled overlap by lightly
perturbing a small subset of clean pairs and retaining them as positives,
and by adding mined hard negatives with unreliable semantics despite
acceptable acoustic quality.
\subsection{Training Pipeline}
\begin{figure*}[t]
\centering
\includegraphics[width=0.93\textwidth]{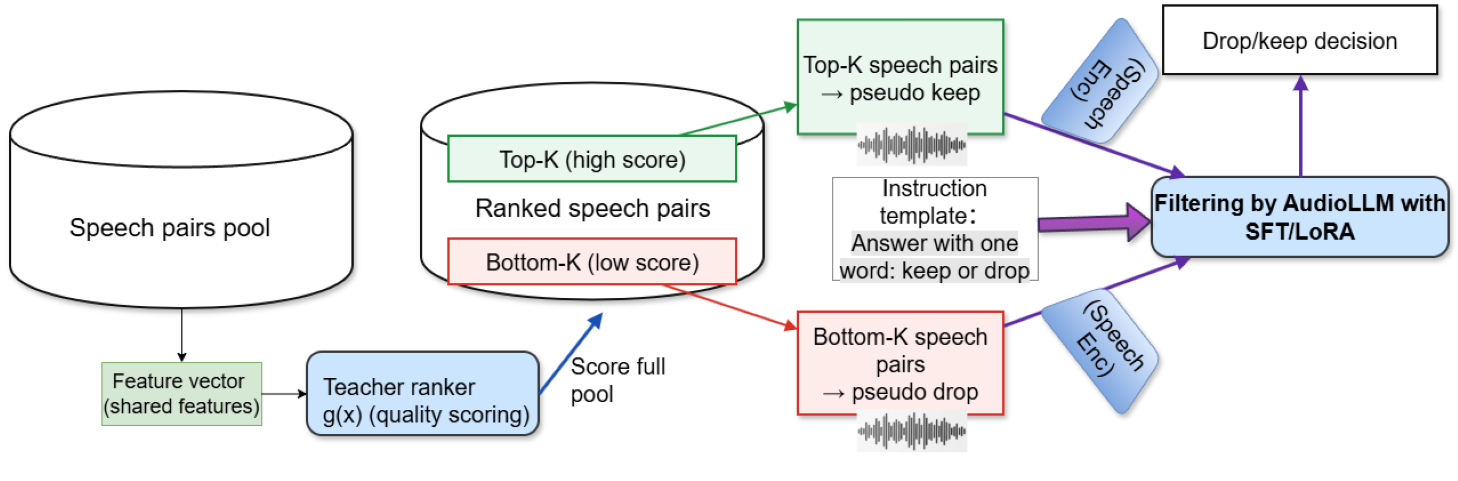}
\caption{
Overview of the proposed two-stage filtering framework.
A lightweight ranker first scores the full speech pair pool and produces pseudo labels by selecting top-K and bottom-K examples.
An audio-language model is then fine-tuned using instruction-following
supervision to directly predict keep/drop decisions from raw speech pairs.
}
\label{fig:pipeline}
\end{figure*}
\subsubsection{Two-stage training pipeline}
We use a two-stage pipeline. In the first stage, a lightweight ranker learns to
score each speech--speech pair using weak supervision signals that combine
acoustic quality, perceptual quality, and semantic consistency. In the second
stage, we use the ranker to score a large unlabeled pool, take the top-ranked
and bottom-ranked examples as pseudo labels, and fine-tune an audio-language
model to predict \texttt{keep}/\texttt{drop} directly from raw audio.The overall filtering framework is illustrated in Figure~\ref{fig:pipeline}.

\subsubsection{Stage I: Ranking model for pseudo-labeling}
\label{stage1}
We start from the constructed training set in
Section~\ref{sec:dataset_generation}, which provides paired examples with weak
quality labels (clean vs.\ noisy). From each speech--speech pair, we extract a
compact set of quality features. We then impose a simple preference assumption:
any clean pair should be ranked above any noisy pair. Based on these labeled
preference pairs, we train a lightweight ranking model as a \emph{teacher} to
predict a continuous quality score for each example, optimized with a standard
pairwise ranking objective:
\[
\mathcal{L}_{\mathrm{rank}}
= - \log \sigma\big( g(\mathbf{x}_i) - g(\mathbf{x}_j) \big).
\]
We finally apply the trained ranker to the full candidate pool and take the
top-ranked and bottom-ranked subsets as high-confidence pseudo labels,
corresponding to \texttt{keep} and \texttt{drop}, respectively.

\subsubsection{Stage II: Fine-tuning an audio-language model}
In Stage~II, we fine-tune an audio-language model (audio LLM) as a
\emph{student} using the pseudo-labeled pairs distilled from the ranker. Each training example consists of paired source/target audio and a discrete target decision (\texttt{keep} or \texttt{drop}), formatted as an instruction-following instance. This teacher--student design
transfers weak, feature-level quality cues into an audio-native filtering model that learns directly from raw speech.
\section{Experiments}
\label{gen_inst}

\subsection{Experiments Setup}
Our experiments adopt the speech-to-unit (S2U) architecture introduced by
\cite{lee2021s2ut} as the backbone speech-to-speech translation model, where target speech is encoded into discrete 
acoustic units and translation is 
performed at the unit level. In this work, we remove the auxiliary task (e.g., source and target
CTC, auto-encoding) in training the S2ST system for simplification.

Following prior work \cite{du2023transspeech}, we conduct all experiments on the CVSS-C corpus \cite{jia2022cvss} (FR$\rightarrow$EN), 
which provides 
high-quality synthetic target speech paired with real human speech on the 
source side. To introduce large-scale supervision, we incorporate SpeechMatrix 
\cite{chen2022speechmatrix}. SpeechMatrix offers wide domain coverage and substantial 
acoustic diversity, making it suitable for evaluating the robustness of 
data filtering methods.

\subsection{Baseline Setup}
The baseline S2ST system is trained on CVSS-C (FR--EN) mixed with
automatically mined SpeechMatrix pairs (FR--EN) without filtering.
Unless otherwise noted, all filtering experiments use CVSS-C plus 20\%
of SpeechMatrix, where filtering is applied only to the mined portion and
CVSS-C is always retained.

% To control data quantity, we include a random-selection baseline by
% keeping 25\%, 50\%, or 75\% of the mined portion (5\%, 10\%,
% or 15\% of the full SpeechMatrix pool), averaged over three seeds.

We evaluate quality-based filtering baselines using automatically computed signals (SNR, MOS, LLM-based semantic similarity, etc.). We also compare our method with the trained ranking model in the stage shown in ~\ref{stage1}.

% Unless otherwise specified, S2ST training hyperparameters follow the Fairseq S2UT recipe.\footnote{\label{fn}\url{https://github.com/facebookresearch/fairseq/blob/main/examples/speech_to_speech/docs/direct_s2st_discrete_units.md}} For each filtering method, the S2ST model is retrained from scratch on the
% selected data with the same training configuration and schedule(4$\times$A100, 300k max updates, FP16, early stopping with patience 5). We use the official CVSS-C dev set for early stopping and report ASR-BLEU (sacreBLEU) on the test set using a fixed wav2vec2 ASR, following the version specified in \hyperref[fn].
Unless otherwise specified, S2ST training hyperparameters follow the Fairseq S2UT recipe\footnote{\url{https://github.com/facebookresearch/fairseq/blob/main/examples/speech_to_speech/docs/direct_s2st_discrete_units.md}}. For each filtering method, the S2ST model is trained from scratch on the selected data with the same training configuration and schedule: 4$\times$A100, 300k max updates, FP16, and early stopping with patience 5. We use the official CVSS-C dev set for early stopping and report ASR-BLEU (sacreBLEU) on the test set using a fixed wav2vec 2.0 ASR specified in the Fairseq S2UT recipe.

\subsection{Implementation of our filtering framework}
For quality signals, the SNR is estimated using the Brouhaha \cite{lavechin2023brouhaha}, the MOS is estimated
 using UTMOS \cite{utmos2022}. For cross-lingual semantic
consistency, we employ two complementary signals. First, we
use Qwen3.1-Instruct (7B) \cite{qwen3technicalreport} as an LLM-based judge over whisper-large-v2 ASR \cite{radford2022whisper} produced transcript to assign a translation adequacy score between
0 and 100 with instruction-style prompt. Second, we translate the source speech transcript with an external machine translation system based on LLaMA-X \cite{lu2024llamax} and compute BLEURT \cite{sellam-etal-2020-bleurt} between
 the MT output and the target transcription to obtain a
text-mediated semantic similarity score. Additional semantic metrics such as BLASER 2.0-QE \cite{dale-costa-jussa-2024-blaser}, which operate directly on speech embeddings
and do not require ASR, are evaluated in ablation experiments.

We construct 15,902 high-confidence positive pairs from SpeechMatrix
using strict thresholds on SNR ($\ge35$), MOS ($\ge2.0$),
LLM adequacy ($\ge90$), and BLEURT ($\ge0.8$), and generate an equal
number of negatives via the degradations described in Section~3. We will release the ranker training data and
augmentation configuration to facilitate reproducibility.

\textbf{Stage I: Ranker and pseudo-labeling.}
We train a lightweight ranking model using automatically computed
signals (SNR, MOS, LLM adequacy, BLEURT). We adopt LambdaMART
\cite{burges2010ranknet2lambdamart} implemented in LightGBM
\cite{ke2017lightgbm} to output a continuous quality score for each sample pair and then score a large unlabeled pool. We use 300 trees (lr=0.05, max depth=6, min leaf size=20, subsample=0.7) and split the data into 10\% dev and 10\% test sets.
The top-$K$ and bottom-$K$ samples are selected as high-confidence
\texttt{keep} and \texttt{drop} pseudo labels. Unless otherwise stated,
we use $K=15\mathrm{k}$.

\textbf{Stage II: Audio LLM fine-tuning.}
Unless otherwise specified, we use
Qwen2-Audio \cite{chu2024qwen2audio} with 4-bit quantization and LoRA
($r{=}16$, $\alpha{=}32$, dropout 0.05). Each sample is formatted as a
chat-style prompt with the two audio inputs and instruction, predicting the
assistant response (\texttt{keep} or \texttt{drop}) using causal LM loss.
We train for 2 epochs (lr=$2{\times}10^{-4}$; batch=8; grad-acc=4) with a 90/10 train–dev split. 
% We also evaluate
% Qwen2.5-Omni-7B \cite{xu2025qwen25omni} and Audio Flamingo 3 \cite{kong2024audioflamingo}
% as alternative architectures.
We also evaluate Audio Flamingo 3 \cite{kong2024audioflamingo}
as alternative architectures

\section{Results}
\subsection{Comparison with Baseline}
Table~\ref{tab:rule} summarizes the comparison between filtering strategies on CVSS-C + SpeechMatrix data. Our finetuned Qwen2-Audio
classifier drops roughly 477k pairs (row 15), and the resulting retained set
achieves 22.72 BLEU, improving by approximately +1.4 BLEU over the
unfiltered baseline (row 3).

% Random inclusion of mined data improves BLEU over the CVSS-C baseline
% (rows 1–2), peaking at 15\% inclusion (row 5) and slightly degrading
% when more data are added (row 6). This indicates substantial quality
% variability in the mined corpus and shows that scaling data alone does
% not reliably improve S2ST performance.

Acoustic and perceptual filtering (rows~4--7) yields
only modest gains and is sensitive to threshold choice, suggesting that
while acoustic quality correlates with usefulness, it is difficult to
exploit with hand-crafted rules. In contrast,
semantic filtering is more effective: a 70B LLM (row 9) and BLEURT
(row 10) achieve stronger performance, confirming that semantic
alignment is a key bottleneck in mined S2ST data.However, the 70B LLM based filtering retains larger
amount of data to reach this performance compared with our method, and they do not explicitly
model speech-conditioned alignment.

To test whether incorporating speech information can further improve
semantic filtering, we conduct a matched-budget comparison at
approximately 477k pairs (rows~11--15). We include strong semantic SOTA
baselines: BLEURT and a 70B LLM (rows 12 and 14), as well as BLASER 2.0-QE
(row 13), which incorporates limited speech
modeling and has been shown effective for filtering. All semantic
methods outperform random selection (row 11). Nevertheless, our audio-LLM filtering (row 15) achieves the best BLEU
(22.72) under the same data budget.

Despite using an 8B audio model, our method surpasses a 70B text-only
LLM, indicating that speech-conditioned semantic modeling is critical
for S2ST data filtering.

\begin{table}[t]
\centering
\small
\setlength{\tabcolsep}{2pt}
\begin{tabular}{r lcc}
\toprule
\textbf{\#} & \textbf{Filtering / Dataset} & \textbf{Retained Pairs} & \textbf{BLEU} \\
\midrule
1 & CVSS-C (FR--EN)\textsuperscript{\dag} & 207{,}365 & 15.44 \\
2 & CVSS-C (reproduced) & 207{,}365 & 15.93 \\

\midrule
% 3 & Random keep 5\% SpeechMatrix & 309{,}090 & 18.59\\
% 4 & Random keep 10\% SpeechMatrix & 410{,}815 & 20.85\\
% 5 & Random keep 15\% SpeechMatrix & 512{,}540 & \textbf{21.64}\\
3 & Keep 20\% SpeechMatrix & 614{,}265 & 21.32\\

\midrule
4 & SNR filtering ($\ge$25 dB) & 553{,}800 & 21.15 \\
5 & SNR filtering ($\ge$30 dB) & 474{,}570 & 21.46 \\
6 & MOS filtering ($\ge$2.0) & 479{,}132 & 21.06 \\
7 & MOS filtering ($\ge$2.2) & 408{,}410 & 20.72 \\
8 & LLM filtering (LLaMA-8B $\ge$80) & 431{,}319 & 19.95 \\
9 & LLM filtering (LLaMA-70B $\ge$80) & 577{,}749 & 22.42 \\
10 & BLEURT filtering ($\ge$0.7)&434{,}254&21.94\\
\midrule
11 & Random keep & 477{,}773 & 21.27\\
12 & BLEURT filtering & 477{,}773 & 22.09 \\
13 & BLASER 2.0-QE filtering & 477{,}773 & 21.71 \\
14 & LLM filtering (70B) & 477{,}773& 22.32 \\
\midrule
\textbf{15} & \textbf{Ours (Audio-LLM filter)} & \textbf{477{,}773} & \textbf{22.72} \\
\midrule
16 & Stage II only  & 477{,}773 & 21.91 \\
17 & Stage II only  & 577{,}749 & 22.49 \\
18 &BLASER 2.0-QE ablation & 482{,}591 & 21.81 \\
19 &BLEURT+LLM ablation  & 469{,}712 & 22.08 \\
20 & Audio Flamingo 3  & 405{,}468 & 21.53 \\

\bottomrule
\end{tabular}
\caption{Filtering comparison on CVSS-C + SpeechMatrix FR$\rightarrow$EN. 
\textsuperscript{\dag}15.44 BLEU reported in \cite{du2023transspeech}.}

\label{tab:rule}
\end{table}
\subsection{Ablation study}
\subsubsection{Effect of removing Stage I.}
First, we evaluate the effects of Stage I. Directly training the audio LLM on clean vs.\ noisy pairs without the ranker removes only $\sim$1.9\% of the mined data (602{,}593/614{,}265; results omitted from Table~\ref{tab:rule} due to negligible filtering), indicating that synthetic
degradations alone cannot capture real noise distributions.

\subsubsection{Effect of removing Stage II.}
We evaluate the effects of Stage II by using the Stage~I ranker alone for data selection.
Selecting the top 477k pairs according to the ranker score (row~16)
yields 21.91 BLEU, already a strong improvement over rule-based and
acoustic baselines. Since the ranker produces continuous scores, we
also increase the retained size to 577k pairs to match the data scale
of the 70B LLM baseline. Under this setting (row~17), performance
further improves to 22.49 BLEU, slightly surpassing the 70B LLM filter
(row~9). These results indicate that the Stage~I ranker learns a
reasonably reliable quality ordering over mined pairs. Nevertheless, the full two-stage
framework still performs best under the same data budget.
\subsubsection{Effect of acoustic and perceptual signals.}
As the quality signals described in~\ref{sec:signals} are optional and
can be freely combined, we study the effect of using semantic signals
only while keeping all other settings unchanged. We consider two
variants: (i) using BLASER 2.0-QE alone (row~18). (ii) a joint usage
of BLEURT and LLM-7B adequacy scores (row~19). Both settings outperform
the unfiltered baseline but remain below the full multi-signal ranker,
indicating that acoustic and perceptual signals provide complementary
information.
\subsubsection{Effect of model choice.}
Replacing Qwen2-Audio with Audio Flamingo~3 (row~20) reduces performance
to 21.53 BLEU. This is likely due to input constraints: the model accepts
only a single audio input, requiring source and target concatenation
with silence.Dual-audio conditioning appears more suitable for S2ST filtering.
\subsubsection{Generalization to German→English (DE--EN)}
% We further evaluate on DE$\rightarrow$EN using the same training
% pipeline and model architecture. For this experiment, we adopt a simplified setup using only
% SNR, MOS, and BLASER~2.0-QE signals.
% The training data consists of full CVSS-C corpus (127{,}822 pairs) and
% 482{,}446 SpeechMatrix pairs.

% Training on unfiltered data yields 12.64 BLEU.
% After filtering 157{,}071 mined pairs with our method, performance improves to 15.04 BLEU (+2.40).
Additionally, we further evaluate on DE$\rightarrow$EN under identical training
settings using a simplified signal set (SNR, MOS, and BLASER~2.0-QE).
Training on 127{,}822 CVSS-C pairs and 482{,}446 SpeechMatrix pairs
yields 13.27 BLEU without filtering and 15.14 BLEU (+1.87) after
removing 157{,}071 mined pairs.
\section{Conclusion}

% We study how to train an audio-language model to make keep/drop
% decisions for speech-to-speech translation (S2ST) data filtering. Using
% a simple Rank→Distill strategy to obtain supervision from noisy speech
% pairs, the resulting model performs speech-conditioned data selection
% directly from raw audio and yields consistent BLEU gains over rule-based and semantic filtering baselines.

% The current formulation treats filtering as a binary keep/drop decision,
% which does not directly allow adjusting the retained data size under a
% fixed budget. Future work will explore more flexible scoring strategies,
% such as using decision probabilities for budget-aware selection, as well
% as incorporating more diverse quality signals and evaluating on larger
% multilingual corpora and more scalable training settings.
We study training an audio-language model for keep/drop decisions in speech-to-speech translation (S2ST) data filtering. Using a simple Rank→Distill strategy to derive supervision from noisy speech pairs, the model performs speech-conditioned selection directly from raw audio and outperforms rule-based baseline. The current binary formulation does not allow flexible control of retained data under a fixed budget. Future work will explore probability-based, budget-aware selection.

\section{Acknowledgments}

{This paper is supported by Project W2531054 of the National Natural Science Foundation of China, and the Program for Guangdong Introducing Innovative and Entrepreneurial Teams.

}

% \ifcameraready
%      The Interspeech 2026 organizers
% \else
%      The authors
% \fi
% would like to thank ISCA and the organizing committees of past Interspeech conferences for their help and for kindly providing the previous version of this template.

\section{Generative AI Use Disclosure}
The authors used generative AI tools only for language editing, polishing, and improving the clarity of the manuscript. The authors reviewed and edited all AI-assisted outputs and take full responsibility for the content of the paper.

% \section{Reference Format}

% {\color{blue}It is ISCA policy that papers submitted to Interspeech should refer to peer-reviewed publications. References to non-peer-reviewed publications (including public repositories such as arXiv, Preprints, and HAL, software, and personal communications) should only be made if there is no peer-reviewed publication available, and should be kept to a minimum.

% References should be in standard IEEE format (follows the IEEEtran format), numbered in order of appearance, for example \cite{Davis80-COP} is cited before \cite{Rabiner89-ATO}. For longer works such as books, provide a single entry for the complete work in the References, then cite specific pages \cite[pp.\ 417--422]{Hastie09-TEO} or a chapter \cite[Chapter 2]{Hastie09-TEO}. Multiple references may be cited in a list \cite{Smith22-XXX, Jones22-XXX}.}

\bibliographystyle{IEEEtran}
\bibliography{mybib}

\end{document}